\documentclass[onecolumn, a4paper, conference]{IEEEtran}
\usepackage[top=2cm, left=2cm, right=2cm, bottom=1.5cm]{geometry}
\usepackage{amsmath}
\DeclareMathOperator*{\argmax}{argmax}
\usepackage{algorithm}
\usepackage{algpseudocode}
\usepackage{amssymb}
\usepackage{dsfont}
\usepackage[dvipsnames]{xcolor}
\usepackage[square]{natbib}
\usepackage[hidelinks, pdftex, pdftitle={An Uncertainty-Aware Pseudo-Label Selection Framework using Regularized Conformal Prediction}, pdfauthor={Matin Moezzi (matin.moezzi@gmail.com)}, pdfsubject={Research Paper}, pdfkeywords={Semi-Supervised Learning, Pseudo-Labeling, Conformal Prediction, Uncertainty Quantification}]{hyperref}
\title{\LARGE{An Uncertainty-Aware Pseudo-Label Selection Framework using Regularized Conformal Prediction}}
\author{\IEEEauthorblockN{Matin Moezzi}
\IEEEauthorblockA{\href{mailto:matin.moezzi@gmail.com}{matin.moezzi@gmail.com}}}
\begin{document}
\maketitle
\begin{abstract}
    Consistency regularization based methods are prevalent semi-supervised learning (SSL) algorithms due to their exceptional performance. However, they mainly depend on domain-specific data augmentations, which are not usable in domains where data augmentations are less practicable. On the other hand, Pseudo-labeling (PL) is a general and domain-agnostic SSL approach that, unlike consistency regularization based methods, does not rely on the domain. PL underperforms due to the erroneous high confidence predictions from poorly calibrated models. This paper proposes an uncertainty-aware pseudo-label selection framework that employs uncertainty sets yielded by the conformal regularization algorithm to fix the poor calibration neural networks, reducing noisy training data. The codes of this work are available at: {\color{cyan}\url{https://github.com/matinmoezzi/ups_conformal_classification}}
\end{abstract}
\section{Introduction}
Much of the recent success in training large, deep neural networks is because of the existence of large labeled datasets. Yet, collecting labeled data is expensive for many learning tasks as it necessarily involves expert knowledge. Semi-supervised learning (SSL) seeks to alleviate the need for labeled data by allowing a model to leverage unlabeled data \citep{mixmatch}.

Consistency regularization based methods are a prevalent semi-supervised learning algorithm due to their exceptional performance. However, they mainly depend on domain-specific data augmentations, which are not usable in domains where data augmentations are less practicable. On the other hand, Pseudo-labeling (PL) is a general and domain-agnostic SSL approach that, unlike consistency regularization based methods, does not rely on the domain. PL underperforms due to the erroneous high confidence predictions from poorly calibrated models.

\citep{ups} argues that conventional pseudo-labeling based methods achieve poor results because poor network calibration produces incorrectly pseudo-labeled samples, leading to noisy training and poor generalization. Since pseudo-labeling has been impactful due to its simplicity, generality, and ease of implementation, \citeauthor{ups} propose an uncertainty-aware pseudo-label selection (UPS) framework, which attempts to maintain these benefits while addressing the calibration issue to improve PL performance drastically. PL leverages the prediction uncertainty to guide the pseudo-label selection procedure.

Recent advances in image classifiers (e.g., CNNs) have yielded high accuracy predictions. However, predicting the most likely label is not the only useful thing in consequential settings such as medical diagnostics. In particular, knowing the uncertainty of the model and considering other less probable labels is necessary for image classification applications where decision-making is crucial. For this purpose, \citeauthor{raps} propose an algorithm that modifies any classifier to output a predictive set, representing the quantified uncertainty of the classifier.

In this paper, we incorporate the uncertainty quantification idea into the UPS framework to reduce noisy training data in the process of pseudo-label selection.

\section{Regularized Adaptive Prediction Set (RAPS)}
In order to quantify the model uncertainty, \citeauthor{raps} proposed an algorithm called RAPS, which modifies the classifier to output a predictive set containing the true label with a user-specified probability. RAPS is based on a data-splitting version of the conformal prediction algorithm, a general algorithm to generate predictive sets and satisfies the coverage property for any predictors. The main RAPS contribution is to add a regularization term to the conformal score.

Formally, for an image classifier with a target label $Y \in \mathcal{Y}=\{1,\ldots,K\}$ and a feature vector $X \in \mathbb{R}^d$, let $\mathcal{C}(x, u, \tau):\mathbb{R}^d \times [0,1] \times \mathbb{R} \rightarrow 2^\mathcal{Y}$ be a set function which generates a predictive set for input $X$. The $\tau$ parameter controls the size of the sets such that $\mathcal{C}(x, u, \tau_1) \subseteq\mathcal{C}(x, u, \tau_2)$ if $\tau_1 \leq \tau_2$.\\
First, RAPS selects the smallest $\tau$, $\tau_{ccal}$, that gives at least $1 - \alpha$ coverage on the conformal calibration set (${U_i} \sim U[0,1]$):
\begin{equation}
    \label{eq:1}
    \hat{\tau}_{\text {ccal }}=\inf \left\{\tau: \frac{\left|\left\{i: Y_{i} \in \mathcal{C}\left(X_{i}, U_{i}, \tau\right)\right\}\right|}{n} \geq \frac{\lceil(n+1)(1-\alpha)\rceil}{n}\right\}
\end{equation}
Secondly, RAPS defines $C(x, u, \tau)$ such that:
\begin{equation}
    \label{eq:2}
    \mathcal{C}^{*}(x, u, \tau):=\{y: \rho_{x}(y)+\hat{\pi}_{x}(y) \cdot u+\underbrace{\lambda \cdot\left(o_{x}(y)-k_{r e g}\right)^{+}}_{\text {regularization }} \leq \tau\}
\end{equation}
Eventually, the predictive set for the input $X$ is give by $\mathcal{C}^*(X, u, \hat{\tau}_{ccal})$[Eq. \ref{eq:2}], which $\hat{\tau}_{ccal}$ is given by Eq. \ref{eq:1}. Algorithm \ref{alg:1} \citep{raps_site} illustrates the pseudo code of RAPS.

\section{Uncertainty-Aware Pseudo-label Selection}
Consider an SSL problem consists of a labeled dataset $D_L=\{(x_i,\mathbf{y}_i)\}_{i=1}^{N_L}$ with $N_L$ samples where $x_i$ is the input vector and $\mathbf{y}_i=[y_1^{(i)},\ldots,y_C^{(i)}] \subseteq \{0,1\}^C$ is the corresponding label with $C$ class categories, and an unlabeled dataset $D_U=\{x^{(i)}\}_{i=1}^{N_U}$ with $N_U$ samples. For the unlabeled samples, pseudo-labels $\mathbf{\tilde{y}}^{(i)}$ are generated. Pseudo-labeling based SSL approaches involve learning a parameterized model $f_\theta$ on the dataset $\tilde{D}=\{(x^{(i)}, \mathbf{\tilde{y}}^{(i)})\}^{N_L+N_U}_{i=1}$, with $\mathbf{\tilde{y}}^{(i)} = \mathbf{y}^{(i)}$ for the $N_L$ labeled samples.

In the conventional pseudo-labeling setting, the model is first trained with labeled data. Then, the model is used to predict labels for unlabeled data. The predicted labels (pseudo-labels) are target classes for unlabeled data as if they were true labels. Finally, the pre-trained model is trained in a supervised fashion with labeled and unlabeled data simultaneously \citep{Lee}.

In order to reduce noisy training data, a subset of pseudo-labels is intelligently selected, which are less noisy for training in each iteration. The high-confidence selection process is based on the network’s output confidence probabilities and selects a subset of pseudo-labels by confidence thresholds. Since the poorly calibrated networks lead an incorrect label to have high confidence, the high-confidence based PL does not have sufficient accuracy.

\citeauthor{ups} concluded that prediction uncertainties can reduce the effects of poor calibration. Thus, they propose an uncertainty-aware pseudo-label selection (UPS) process, which selects a more accurate subset of pseudo-labels used in training by employing both high confidence and uncertainty prediction (Eq \ref{eq:4}). \citep{ups} used the MC-Dropout sampling as the uncertainty estimation method.

UPS uses hard pseudo-labeling in which pseudo-labels are obtained directly from the network prediction. Let $\mathbf{p}^{(i)}$ be the probability outputs of a trained network on the sample $x(i)$, such that $p_c^{(i)}$ represents the probability of class $c$ being present in the sample. the hard pseudo-label can be generated for $x^{(i)}$ as:
\begin{equation}
    \label{eq:3}
    \tilde{y}^{(i)}_c=\mathds{1}[p_c^{(i)}\geq \gamma]
\end{equation}

Formally, Let $\mathbf{g}^{(i)}=[g_1^{(i)},\ldots,g_C^{(i)}]\subseteq \{0,1\}^C$ be a binary vector representing the selected pseudo-labels in sample $i$, where $g_c^{(i)} =1$ when ${\tilde{y}}^{(i)}_c$ is selected and $g_c^{(i)} =0$ when ${\tilde{y}}^{(i)}_c$ is not selected. UPS improves the high-confidence selection process such that:
% \begin{equation}
%     \label{eq:4}
%     g_{c}^{(i)}=\mathds{1}\left[p_{c}^{(i)} \geq \tau_{p}\right]+\mathds{1}\left[p_{c}^{(i)} \leq \tau_{n}\right]
%     \rlap{\qquad\text{\small{(High-confidence selection)}}}
% \end{equation}
\begin{equation}
    \label{eq:4}
    g_{c}^{(i)}=\mathds{1}\left[u\left(p_{c}^{(i)}\right) \leq \kappa_{p}\right] \mathds{1}\left[p_{c}^{(i)} \geq \tau_{p}\right]+\mathds{1}\left[u\left(p_{c}^{(i)}\right) \leq \kappa_{n}\right] \mathds{1}\left[p_{c}^{(i)} \leq \tau_{n}\right]
\end{equation}
where $u(p)$ is the uncertainty of a prediction $p$, and $\kappa_p$ and $\kappa_n$ are the uncertainty thresholds, and $\tau_p$ and $\tau_n$ are the confidence thresholds for positive and negative labels.

In each iteration, the parameterized neural network $f_\theta$ is trained on labeled data $D_L$. Then, $f_\theta$ predicts the probability outputs for all unlabeled data $D_U$. Pseudo-labels are created by Eq. \ref{eq:3} and a subset of pseudo-labels are selected following Eq. \ref{eq:4}. Next, $f_\theta$ is trained on a selected subset of pseudo-labels. Loss functions for positive and negative labels are calculated by cross-entropy loss which are given by
\begin{equation}
    \label{eq:5}
    L_{\mathrm{NCE}}\left(\tilde{\boldsymbol{y}}^{(i)}, \hat{\boldsymbol{y}}^{(i)}, \boldsymbol{g}^{(i)}\right)=-\frac{1}{s^{(i)}} \sum_{c=1}^{C} g_{c}^{(i)}\left(1-\tilde{y}_{c}^{(i)}\right) \log \left(1-\hat{y}_{c}^{(i)}\right),
\end{equation}
\begin{equation}
    \label{eq:6}
    L_{\mathrm{BCE}}\left(\tilde{\boldsymbol{y}}^{(i)}, \hat{\boldsymbol{y}}^{(i)}, \boldsymbol{g}^{(i)}\right)=-\frac{1}{s^{(i)}} \sum_{c=1}^{C} g_{c}^{(i)}\left[\tilde{y}_{c}^{(i)} \log \left(\hat{y}_{c}^{(i)}\right)+\left(1-\tilde{y}_{c}^{(i)}\right) \log \left(1-\hat{y}_{c}^{(i)}\right)\right]
\end{equation}
Where $\hat{\mathbf{y}}^{(i)} = f_\theta(x^{(i)})$  and $s^{(i)}=\sum_cg_c^{(i)}$ is the number of selected pseudo-labels for sample $i$. After calculating the losses of selected pseudo-labels, the back propagation step of $f_\theta$ is performed using the sum of labeled and selected pseudo-labels losses.
\section{Our Method}
In this paper, we employ the RAPS framework in the UPS algorithm. RAPS takes an image classifier and outputs a predictive set for each input which represents the uncertainty of the classifier. We use RAPS as a wrapper around the UPS pre-trained model. To be specific, we use the predictive set yielded by RAPS instead of hard pseudo-labeling to predict labels for unlabeled data.

Formally, consider we have $C$ class categories. For input $x^{(i)}$, RAPS produces a predictive set such that $$\mathcal{C}^*(x^{(i)}, u, \hat{\tau}_{ccal}) \subseteq \mathcal{Y}=\{1,\ldots,C\}.$$
Which means for every $c$ in $\mathcal{C}^*(x^{(i)})$, input $x^{(i)}$ belongs to the $c$ class. Consider unlabeled input $\tilde{x}^{(i)}$, for multi-label case, the proposed method predicts the corresponding pseudo-label such that
\begin{align}
    \mathbf{\tilde{y}}^{(i)} & =\{\tilde{y}_1^{(i)},\ldots, \tilde{y}_C^{(i)}\}                                                         \\
    \tilde{y}_c^{(i)}        & =\mathds{1}[c \in \mathcal{C}^*(\tilde{x}^{(i)}, u, \hat{\tau}_{ccal})] \;\;\; \forall c \in \mathcal{Y}
    \label{eq:7}
\end{align}
And for single-label case, the label with the highest conformal score is assigned as the pseudo-label such that:
\begin{align}
    \label{eq:9}
    \tilde{y}_c^{(i)}=\left\{\begin{array}{ll}
        1, & \text{if } c \in \mathcal{C}^*(\tilde{x}^{(i)}, u, \tau_{ccal}) \text{ and } \argmax\limits_{j\in\mathcal{Y}} s_{i,j} = c \\
        0, & \text{Otherwise}
    \end{array}\right.
\end{align}
The conformal score of sample $i$ and class $j$, $s_{i,j}$, is the softmax outputs of the network.

% A high value of uncertainty scalar, $u(X)$, means the model has high uncertainty. Thus, we can interpret the size of the, representing the classifier’s uncertainty predictive set, $\mathcal{C}^*$, as the uncertainty prediction of the model. As a result, we apply the size of the predictive sets as uncertainty estimation in the pseudo-label selection procedure (Eq. \ref{eq:4}), instead of variance-based uncertainty estimation (MC-Dropout). The new selection procedure is given by:

\section{Learning Algorithm}
First, a neural network $f_{\theta, 0}$ is trained on the labeled dataset $D_L$. Once trained, RAPS takes this pre-trained network ($f_{\theta, 0})$ and predicts the probabilities estimation beside predictive sets for all unlabeled data $D_U$. Next, pseudo-labels are created from predictive sets following Eq. \ref{eq:7} (Eq. \ref{eq:9} for the single-label case). Then, a subset of these pseudo-labels is selected using UPS (Eq. \ref{eq:4}). Afterward, another network $f_{\theta, 1}$ is trained on both labeled data $D_L$ and selected pseudo-labels. This procedure is repeated iteratively until the number of selected pseudo-labels converges. A new network is initialized to prevent the error propagation issue in each iteration. This algorithm is illustrated in Algorithm \ref{alg:2}.

\begin{algorithm}[h]
    \caption{RAPS}
    \begin{algorithmic}[1]
        \Procedure{RAPS}{the calibration dataset, the model, the new image}
        \State \textbf{calibrate:} perform Platt scaling on the model using the calibration set.
        \State \textbf{get conformal scores:} For each image in the training set, define $E_j=\sum_{i=1}^{k'}(\hat{\pi}_{(i)}(x_j)+\lambda \mathds{1}[j > k_{reg}])$ where $k'$ is the model's ranking of the true class $y_j$, and $\hat{\pi}_{(i)}(x_j)$ is the $i^{th}$ largest score for the $j^{th}$ image.
        \State \textbf{find the threshold:} assign $\hat{\tau}_{ccal}$ to the $1-\alpha$ quantile of the $E_j$.
        \State \textbf{predict:} Output the $k^*$ highest-score classes, where $\sum_{i=1}^{k^*}(\hat{\pi}_{(i)}(x_{n+1})+\lambda \mathds{1}[j > k_{reg}]) \geq \hat{\tau}_{ccal}$.
        \EndProcedure
    \end{algorithmic}
    \label{alg:1}
\end{algorithm}
\begin{algorithm}[h]
    \caption{UPS algorithm using RAPS's prediction sets}
    \label{alg:2}
    \begin{algorithmic}[1]
        \Procedure{UPS-RAPS}{labeled dataset $D_L$, unlabeled dataset $D_U$, $\kappa_p$, $\kappa_n$, $\tau_p$, $\tau_n$, uncertainty estimator $u$}
        \State Train $f_{\theta, 0}$ on $D_L$
        \State $\tilde{D}$ $\leftarrow$ $D_L$
        \For{$k=1\ldots MaxIterations$}\Comment{Repeats until convergence}
        \For {$\tilde{x}^{(i)}$ in $D_U$}
        \State $\mathbf{p}^{(i)}$, predictive set $\mathcal{C}^*$ $\leftarrow$ RAPS($\tilde{D}$, $f_{\theta, i-1}$, $\tilde{x}^{(i)})$\Comment{RAPS from Algorithm \ref{alg:1}}
        \For{$c$ in $\{1,\ldots,C\}$}\Comment{Equation \ref{eq:7}}
        \If{$c$ in $\mathcal{C}^*$}
        \State $\tilde{y}^{(i)}_c$ $\leftarrow$ $1$
        \Else
        \State $\tilde{y}^{(i)}_c$ $\leftarrow$ $0$
        \EndIf
        \State positive mask $\leftarrow$ $\left(u(p_{c}^{(i)}) \leq \kappa_{p} \times p_{c}^{(i)} \geq \tau_{p}\right)\times 1$
        \State negative mask $\leftarrow$ $\left(u(p_{c}^{(i)}) \leq \kappa_{n} \times p_{c}^{(i)} \geq \tau_{n}\right)\times 1$
        \State $g_c^{(i)}$ $\leftarrow$ positive mask + negative mask\Comment{Equation \ref{eq:4}}
        \EndFor
        \State $\tilde{\mathbf{y}}^{(i)}$ $\leftarrow$ $[\tilde{y}^{(i)}_1,\ldots,\tilde{y}^{(i)}_C]$
        \State ${\mathbf{g}}^{(i)}$ $\leftarrow$ $[{g}^{(i)}_1,\ldots,{g}^{(i)}_C]$
        \EndFor
        \State $D_{selected}$ $\leftarrow$ $\{(\tilde{x}^{(i)}, \mathbf{\tilde{y}}^{(i)}, \mathbf{g}^{(i)})\}_{i=1}^{N_U}$\Comment{Selected pseudo-labels}
        \State $\tilde{D}$ $\leftarrow$ $D_L \cup D_{selected}$
        \State Initialize a new network $f_{\theta, i}$
        \State Train $f_{\theta, i}$ using samples from $\tilde{D}$\Comment{Using loss functions in Eq. \ref{eq:5}-\ref{eq:6}}
        \State $f_\theta$ $\leftarrow$ $f_{\theta, i}$
        \EndFor
        \State \textbf{return} $f_\theta$
        \EndProcedure
    \end{algorithmic}
\end{algorithm}
\section{Discussion}
The hard pseudo-labeling method for generating pseudo-labels is based on the network output probabilities, which, as claimed in the original paper, the network suffers poor calibration. Hence, using conformal prediction and predictive sets rather than hard pseudo-labels improves the performance and alleviates the noisy pseudo-labels for training.

\section{Future Work}
The RAPS algorithm and generally the conformal prediction method can adapt with uncertainty estimation methods and modify the conformal score. In particular, predictive sets can be obtained by conformal scores, which are based on variance-based uncertainty scalar or any uncertainty scalar $u(X)$ \citep{gentle}.

\bibliographystyle{abbrvnat}
\bibliography{references.bib}
\end{document}